\pdfoutput=1

\documentclass[11pt]{article}

\usepackage{emnlp2021}

\usepackage{times}
\usepackage{latexsym}

\usepackage[T1]{fontenc}

\usepackage[utf8]{inputenc}

\usepackage{microtype}

\usepackage{amsfonts}       
\usepackage{amsmath}
\usepackage{amssymb}
\usepackage{graphicx}
\usepackage{enumitem}
\usepackage{array}
\usepackage{breakcites}
\usepackage{multirow}
\usepackage{url}

\usepackage{arydshln}
\usepackage{algorithm}          
\usepackage{algpseudocode}      
\usepackage{amsmath}            
\usepackage{amssymb}            

\newcommand{\mranews}{\textsc{MiRAnews}}
\newcommand{\lead}{\textsc{Lead}}
\newcommand{\oracle}{\textsc{EO}}
\newcommand{\pz}{\phantom{0}}

\title{\mranews{}: Dataset and Benchmarks for Multi-Resource-Assisted News Summarization}

\usepackage{xcolor}
\usepackage[normalem]{ulem}

\def\ODdel#1{\bgroup\markoverwith{\textcolor{cyan!80!yellow!80!black!100}{\rule[0.4ex]{2pt}{3pt}}}\ULon{#1}}

\def\IKdel#1{\bgroup\markoverwith{\textcolor{green!80!black!100}{\rule[0.4ex]{2pt}{3pt}}}\ULon{#1}}
\def\VR#1{{\color{magenta}VR: \it #1}}
\def\VRdel#1{\bgroup\markoverwith{\textcolor{gray}{\rule[0.5ex]{2pt}{1pt}}}\ULon{#1}}

\def\XXdel#1{\bgroup\markoverwith{\textcolor{green!10!orange!90!}{\rule[0.5ex]{2pt}{1pt}}}\ULon{#1}}

\newcommand{\ignore}[1]{}
  
\author{Xinnuo Xu$^1$, Ond\v{r}ej Du\v{s}ek$^2$, Shashi Narayan$^3$, Verena Rieser$^1$ and Ioannis Konstas$^1$ \\
  $^1$The Interaction Lab, MACS, Heriot-Watt University, Edinburgh, UK \\
  $^2$Charles University, Faculty of Mathematics and Physics, Prague, Czechia \\ 
  $^3$Google Research\\
  {\tt \{xx6, v.t.rieser, i.konstas\}@hw.ac.uk} \\
  {\tt odusek@ufal.mff.cuni.cz} \\
  {\tt shashinarayan@google.com} \\}

\begin{document}
\maketitle
\begin{abstract}
One of the most challenging aspects of current single-document news summarization
is that the summary often contains `extrinsic hallucinations', i.e., facts that are not present in the source document, which are often derived via world knowledge.
This causes summarization systems to act more like open-ended language models tending to hallucinate facts that are erroneous.
In this paper, we mitigate this problem with the help of multiple supplementary resource documents assisting the task.
We present a new dataset \mranews~ 
and benchmark existing 
summarization models.\footnote{Our code and data are available at:\\ \url{https://github.com/XinnuoXu/MiRANews}}
In contrast to multi-document summarization, which addresses multiple events from several source documents, we still aim at generating a summary for a single document.
We show via data analysis that it's not only the models which are to blame: more than 27\% of facts mentioned in the gold summaries of \mranews~are better grounded on assisting documents than in the main source articles. 
An error analysis of generated summaries from pretrained models fine-tuned on \mranews{}~reveals that this has an even bigger effects on models:
assisted summarization reduces
55\% of  hallucinations when compared to single-document summarization models trained on the main article only. 
\end{abstract}

\section{Introduction}

\begin{figure}[tb]
  \includegraphics[width=1\columnwidth]{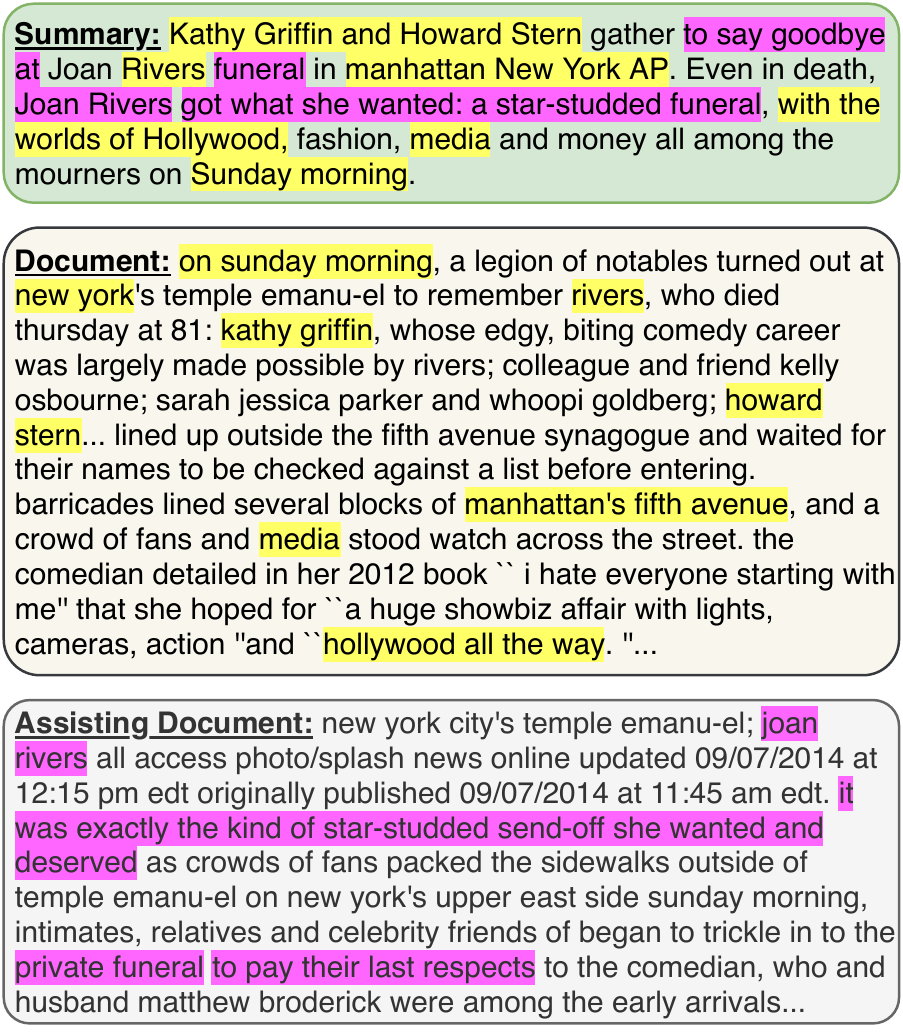}
  \caption{An example where the summary (top section) contains information that is not explicitly included in its main document (middle section), but is covered in the related assisting document (bottom section). We highlight the information in the summary that is aligned to its corresponding main and assisting documents with yellow and pink colors, respectively.}
  \label{fig:dataset_ex}
\end{figure}
 
\begin{figure*}
    \centering\includegraphics[width=1.0\textwidth]{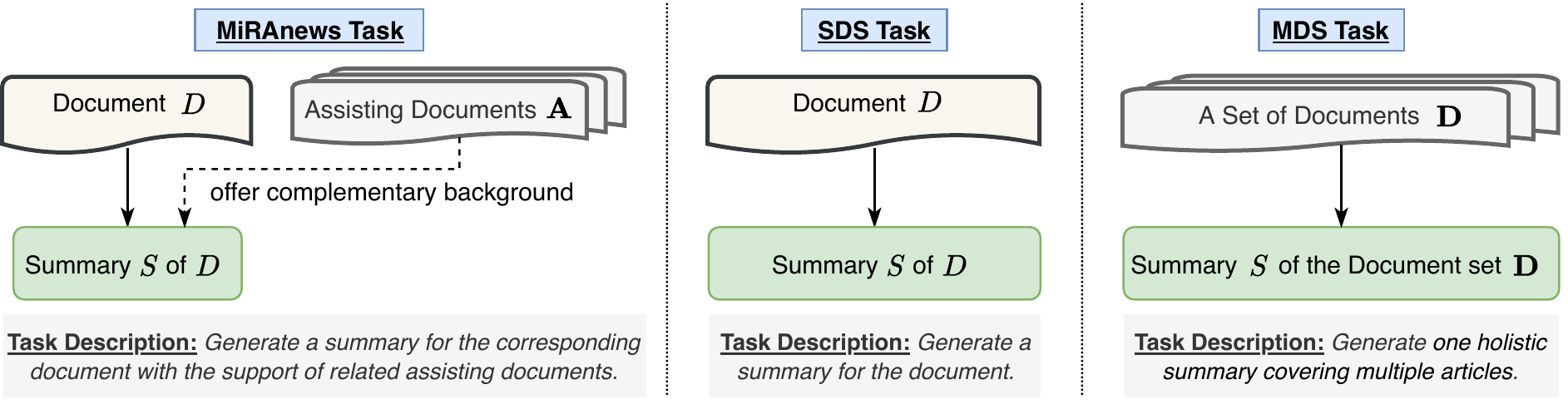}
    \caption{Comparison of Summarization tasks.
    Single-document Summarization (SDS task) focuses on generating summary $S$ based on a single document $D$.  Multi-document Summarization (MDS task) creates a holistic summary $S$ covering multiple articles $\mathbf{D}$. The \mranews{} task differs by producing summary $S$ based only on the events pertinent in the \textit{main} article $D$, while reaching to a set of assisting documents $\mathbf{A}$ for complementary background.}
    \label{fig:mra_task}
\end{figure*}

The vast majority of current research on abstractive summarization is aimed at single-document news summarization due to the widespread availability of data, e.g.\
(NY Times; \citet{sandhaus2008new}, 
CNN/DailyMail; \citet{10.5555/2969239.2969428}, 
Newsroom;  \citet{grusky-etal-2018-newsroom}, 
XSum; \citet{narayan-etal-2018-dont}, 
MLSUM; \citealt{scialom-etal-2020-mlsum}).
The datasets are 
curated by pairing a single document with human authored highlights/description as the summary.
This task is typically approached using 
 conditional generation models, including sequence-to-sequence architectures with attention and copy mechanisms \cite{46111}, 
 Transformers \cite{liu-lapata-2019-hierarchical}, 
and pre-trained language modeling \cite[e.g.][]{Radford2019LanguageMA,lewis-etal-2020-bart}.

While these SotA summarization models reach a high level of fluency and coherence, they are also highly prone to hallucinating content that is 
not grounded by the input document.
\citet{maynez-etal-2020-faithfulness} classified hallucinations into {\bf {\em intrinsic}} that mistakenly 
manipulate information from the source document resulting in \textit{counterfactual} output,
and {\bf {\em extrinsic}} that introduce information not 
grounded in the document (see Figure \ref{fig:dataset_ex}). Extrinsic hallucinations are further broken down into `factual', i.e., holding true in real life, and `counterfactual'.



Similar to \cite{maynez-etal-2020-faithfulness}, we find not only the models are to blame, but also the datasets: human-written summaries contain up to 36\% external facts which are not \textit{faithful}, i.e., covered by the single input document. In other words: the summaries also contain `extrinsic hallucinations'. Moreover, facts which are present are often re-phrased or shortened in the summary in ways which requires world knowledge.
 Consider the example in Figure~\ref{fig:dataset_ex}, where 
 the surname ``{\em Rivers}'' used throughout the document (middle section), is elaborated as the full name ``{\em Joan Rivers}'' in the summary (top section), i.e.\ adding information. Meanwhile, ``{\em celebrities lined up outside the fifth avenue synagogue}'' in the document is specified as ``{\em say goodbye at Joan Rivers funeral}'' in the summary, which requires world knowledge. Moreover, the fact about an ``a star-studded funeral'' is not mentioned explicitly in the document. 
 Any summarization model that is agnostic to such data divergence issues between the source and target texts \cite{dhingra-etal-2019-handling} will function more as an open-ended language model and will be prone to extrinsic hallucinations.

In this work, we tackle the problem of \textit{extrinsic hallucinations} by introducing a new task, Multi-Resource-Assisted News 
Summarization and a novel dataset (\mranews). 
Following \citet{maynez-etal-2020-faithfulness}, we regard the incorporation of background knowledge within a generated summary as the desired property. 
However, instead of sourcing this knowledge via pretraining on large datasets,\footnote{Although they report \textsc{BertS2S} \cite{rothe-etal-2020-leveraging} to output more factual hallucinations in the summary than their non-pre-trained counterparts on XSum \cite{narayan-etal-2018-dont}, still over 90\% of the total hallucinations are incorrect.} 
we base our work on the assumption that articles from alternative news resources 
covering the same news event can complement the background knowledge in an easier to learn, more direct, and explainable way.
Consider the example in Figure~\ref{fig:dataset_ex}, where the assisting document (bottom section) from another news resource recounts some facts in the summary (highlighted in pink) in a more explicit way. 

Note that, as shown in Figure~\ref{fig:mra_task} (left), {\bf our task is different from both  Single-document Summarization} (SDS, middle) {\bf and  Multi-document Summarization} (MDS, right): 
SDS aims at generating a summary for a single main document, while we aim to generate a target summary $S$ for a single document $D$ with supporting facts from multiple assisting documents $\mathbf{A}$.
In this paper:

\noindent$\bullet$ We introduce a new task, Multi-Resource-Assisted News summarization, aiming at generating a summary for the corresponding news article with the support of related assisting documents.

\noindent$\bullet$ We create and release a new dataset (\mranews) introducing a novel automatic data collection method which gathers multiple assisting news articles from different news resources for a document-summary pair. 

\noindent$\bullet$ We introduce new \textit{referenceless} metrics, which quantitatively evaluate extrinsic hallucinations both in summarization datasets and output summaries, and confirm that introducing assisting documents offers better grounding to more than 27\% of facts mentioned in the reference summaries.

\noindent$\bullet$ We report benchmark results using models both fine-tuned and trained from scratch on \mranews{}. We show that modeling assisting documents effectively introduces external facts in the summaries that are grounded on the assisting documents, resulting in 55\%  less counterfactual hallucinations than SDS systems. 


\section{Data Collection}

\begin{figure*}[tb]
  \includegraphics[width=1.0\textwidth]{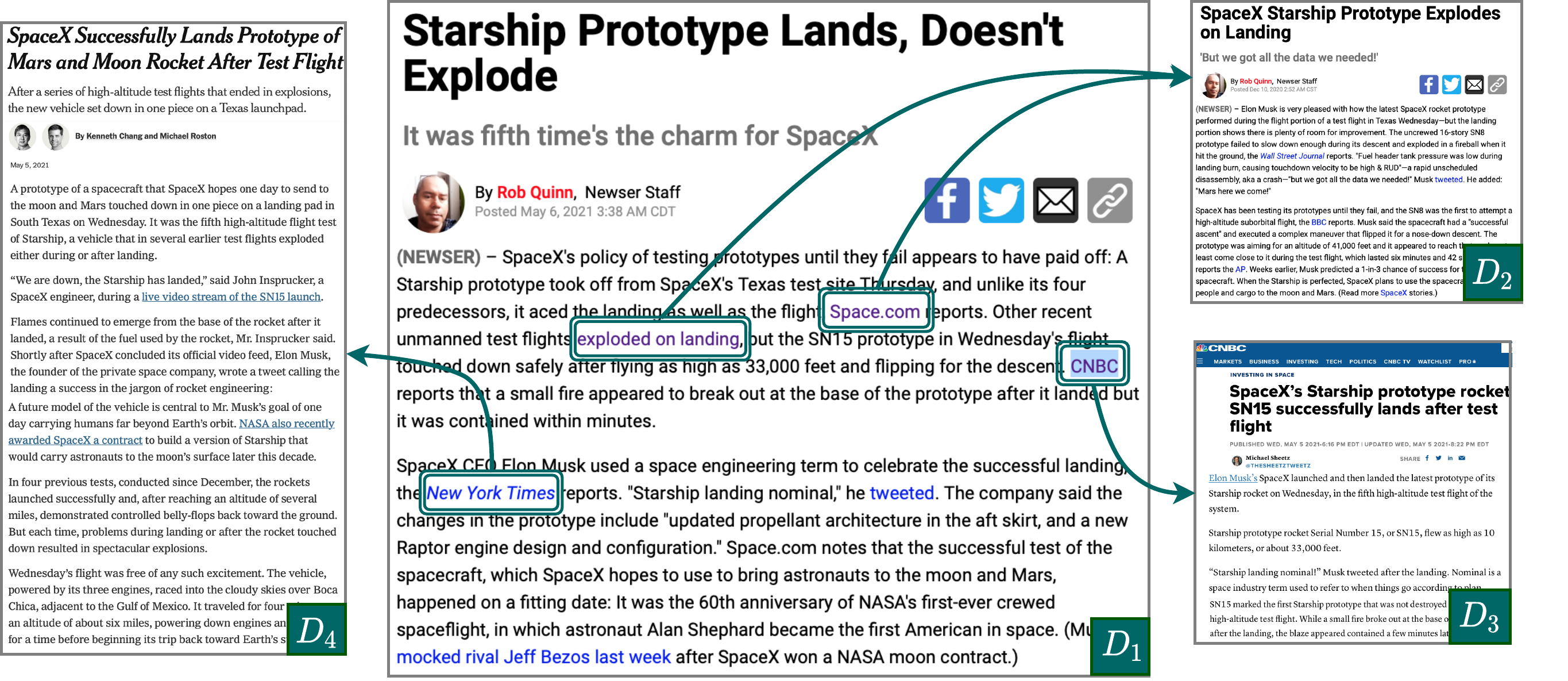}
  \caption{Example of a page on
  {\em newser.com}: a \url{newser.com} article is a news event including editor-picked links to relevant news articles from other news websites. This example shows the webpage \url{https://www.newser.com/story/305823/starship-prototype-lands-doesnt-explode.html}.
  In the webpage ($D_1$), three extra news pieces  ($D_2$, $D_3$, $D_4$) from {\em nytimes}, {\em newser}, and {\em CNBC} are linked. All of these four news articles report on the same event of  starship prototype landing.
  }
  \label{fig:newser}
\end{figure*}

\ignore{
We automatically create a dataset containing news document-summary pairs. Each pair is accompanied by a set of assisting documents. \VR{Repetition? Can be deleted to safe space?}
}

\paragraph{Data Resource.} 

Following \citet{fabbri-etal-2019-multi}'s MDS efforts, we use the news aggregation portal
\url{newser.com} 
to collect news articles with their assisting documents,
where 
each webpage 
 reports on a news event and includes editor-picked links to the relevant news articles from other news websites. 
An example 
is in Figure~\ref{fig:newser}: three news articles ($D_2$, $D_3$, $D_4$) from {\em nytimes}, {\em newser}, and {\em CNBC} are linked to the webpage ($D_1$), all of which report on the same event of starship prototype landing.

\paragraph{News Cluster and Content Extraction.} 
We consider each article on \url{newser.com}, together with the pages cited therein, as a cluster about one news event. 
We extract the document and the corresponding summary from each webpage automatically, following the method introduced in \textsc{Newsroom} \cite{grusky-etal-2018-newsroom}.\footnote{We use the data scraping and data extraction code from \url{https://github.com/lil-lab/newsroom}.}
Specifically,  
the documents are constructed from the HTML main text body excluding HTML markups, inline advertising, images/videos, and captions,
while the target summary $S$ is extracted from the document's metadata fields, e.g.\ {\em og:description}, {\em twitter:description}, {\em description},  
which are often written by editors and journalists to appear on social media services or as search engine webpage descriptions. 
Hence, for each cluster $\mathbf{C}$, we collect paired documents and summaries $\mathbf{C}=\left \{ \left (  D_1, S_1 \right ), \left (  D_2, S_2 \right ) \cdots \left (  D_m, S_m \right ) \right \}$, where $m$ is the number of webpages in the cluster.

\paragraph{Collecting Assisting Documents.}
We first represent all documents in the news cluster $\mathbf{C}$ as $\mathbf{D} = \left \{ D_1, D_2\cdots D_m \right \}$. 
In turn, we take $\mathbf{A}_i=\mathbf{D}-D_i$ as the assisting documents for each document $D_i$ and its summary $S_i$ in the cluster.
Thus, for a cluster including $m$ corresponding webpages, we create $m$ examples. 
Each of them contains one document, its summary, and $m-1$ assisting documents, denoted as $\left ( D_i, S_i, \textbf{A}_i \right )$.

Accordingly, we create the full \mranews{} dataset $\mathcal{D} = \left \{ \left ( D_i, S_i, \textbf{A}_i \right ) \right \}_{i=1}^M$ by collecting examples from all available 57K \url{newser.com} pages 
following \citet{fabbri-etal-2019-multi}. 
Note that, before creating the clusters,
we first randomly split the webpages 
into training (80\%), validation (10\%), and test (10\%) set, and then generate examples within each set in order to prevent data leaking, i.e.\ each document is only included in one of the sections (regardless of main/assisting role).

\section{Data Analysis} \label{sec:data_analysis}

\begin{table*}[tb]
  \centering\small
  \begin{tabular}{l|rrr|cc|cc|cc}
  \hline
  \multicolumn{1}{c|}{\multirow{2}{*}{{\bf Datasets}}} & \multicolumn{3}{c|}{{\bf \# examples}} & \multicolumn{2}{c|}{{\bf avg. doc len}} & \multicolumn{2}{c|}{{\bf avg. summ len}} & \multicolumn{2}{c}{{\bf vocabulary size}}  \\ 
  & \multicolumn{1}{c}{\bf train} & \multicolumn{1}{c}{\bf valid} & \multicolumn{1}{c|}{\bf test} & {\bf words} & {\bf sents} & {\bf words} & {\bf sents} & {\bf document}  & {\bf summary} \\ \hline \hline
  CNN & 90,266 & 1,220 & 1,093 & 760.50 & 33.98 & 45.70 & 3.59 & 343,516 & 89,051 \\
  DailyMail & 196,961 & 12,148 & 10,397 & 653.33 & 29.33 & 54.65 & 3.86 & 563,663  & 179,966 \\
  NY Times & 589,284 & 32,736 & 32,739 & 800.04 & 35.55 & 45.54 & 2.44 & 1,399,358 & 294,011 \\
  XSum & 204,045 & 11,332 & 11,334 & 431.07 & 19.77 & 23.26 & 1.00 & 399,147 & 81,092 \\
  Newsroom & 995,041 & 105,760 & 105,760 & 658.60 & --- & 26.70  & --- & \multicolumn{2}{c}{6,925,712} \\ \hdashline
  MiRAnews & 119,150 & 13,018 & 15,670 & 690.20 & 32.82 & 33.24 & 1.81 & 736,496 & 136,304 \\ \hline
  \end{tabular}
  \caption{Comparison of summarization datasets: size of training, validation, and test set, average document (source) and summary (target) length (in terms of words and sentences), and vocabulary size for both source and target. The numbers for CNN DailyMail, NY Times, and XSum are reported in \citet{narayan-etal-2018-dont}. The numbers for Newsroom are reported in \citet{grusky-etal-2018-newsroom}. All tokens in \mranews{} vocabulary are lowercased.
  \label{table:data_length}}
\end{table*}

\mranews{} contains 150K examples in total, with an average of 1.7 assisting documents per instance
\footnote{The minimum and maximum number of assisting documents in each example is 1 and 4. We keep the four assisting documents at most for each example.
} Table~\ref{table:data_length} compares \mranews{} with popular large scale summarization datasets. 
\mranews{} is similar in size to CNN; 
document and summary average lengths in \mranews{} are similar to CNN, DailyMail \cite{10.5555/2969239.2969428}, NY Times \cite{sandhaus2008new}, and Newsroom \cite{grusky-etal-2018-newsroom}, but longer than XSum \cite{narayan-etal-2018-dont}. 

\begin{table*}[tb]
  \centering\small
  \begin{tabular}{l|cccc|ccc|ccc} 
  \hline
  \multirow{2}{*}{{\bf Dataset}} & \multicolumn{4}{c|}{{\bf \% of novel n-grams in gold summary}} & \multicolumn{3}{c|}{{\textbf{\lead}}} & \multicolumn{3}{c}{{\textbf{\oracle}}}  \\
    & {\bf 1-gram} & {\bf 2-gram} & {\bf 3-gram} & {\bf 4-gram} & {\bf R1} & {\bf R2} & {\bf RL} & {\bf R1} & {\bf R2} & {\bf RL} \\ \hline \hline
  CNN & 16.75 & 54.33 & 72.42 & 80.37 & 29.15 & 11.13 & 25.95 & 50.38 & 28.55 & 46.58 \\
  DailyMail & 17.03 & 53.78 & 72.14 & 80.28 & 40.68 & 18.36 & 37.25 & 55.12 & 30.55 & 51.24 \\
  NY Times & 22.64 & 55.59 & 71.93 & 80.16 & 31.85 & 15.86 & 23.75 & 52.08 & 31.59 & 46.72 \\
  XSum & 35.76 & 83.45 & 95.50 & 98.49 & 16.30 & \pz1.61 & 11.95 & 29.79 & \pz8.81 & 22.65 \\ \hdashline
  MiRA(S-D) & 16.31 & 35.43 & 42.72 & 45.75 & 38.38 & 28.78 & 34.24 & 59.38 & 47.71 & 53.18 \\
  MiRA(S-A) & 32.11 & 75.90 & 90.62 & 94.96 & 18.32 & \pz4.10 & 12.35 & 34.42 & 12.76 & 23.33 \\ \hline
  MiRA(S-D\&A) & 10.29 & 30.36 & 40.01 & 44.04 & --- & --- & --- & 61.36 & 49.18 & 54.47 \\\hline
  \end{tabular}
  \caption{Corpus bias towards extractive methods in popular dataset 
  and \mranews{}. We show the proportion of novel n-grams in gold summaries. We also report ROUGE scores for the LEAD baseline and the extractive oracle system EXT-ORACLE. Results are computed on the test set.
  The numbers for CNN, DailyMail, NY Times and XSum are reported by \citet{narayan-etal-2018-dont}. For \mranews{}, S-D, S-A and S-D\&A represent summary-document, summary-assisting document and summary-document \& assisting document, respectively.  \label{table:data_novelty}}
\end{table*}


\subsection{Bias towards Extractive Methods}\label{subsec:ext_bias}

\paragraph{N-gram novelty.} We evaluate the dataset bias towards extractive methods 
 using n-gram novelty introduced in \cite{narayan-etal-2018-dont}. 
This metric reports the percentage of novel n-grams in the gold summaries that do not appear in their source documents. 
Lower values indicate that more n-grams of the summaries appear in the documents, i.e.\ there is more overlapping information that supports the summary, 
leading 
to more extractive summaries.

The left section in Table~\ref{table:data_novelty} shows the results 
in comparison with other commonly used datasets.
\mranews{}(S-D), i.e.\ the percentage of novel n-grams in the summaries $S$ that do not appear in their main document $D$, is lower than in other benchmarks. 
This means that \mranews{}, 
when treated as a SDS task, will benefit extractive methods. 
On the other hand, \mranews{}(S-A), i.e.\ the n-grams novelty of the summaries with respect to their assisting documents $\mathbf{A}$, is much higher, 
comparable with XSum.
this shows that assisting documents in \mranews{} are not redundant to the main documents.
The level comparable to XSum suggests that they indeed describe the same news event, i.e., are relevant to the summaries.

\paragraph{\lead{} and \oracle.} We further evaluate two well established extractive methods on \mranews{} and other benchmarks. 
\lead{} is often used as a strong lower bound for summarization \cite{10.5555/1619499.1619564} and creates a summary by selecting the first few sentences or words in the document.
For \mbox{\mranews{}(S-D)}, we select the first three sentences in the main document, and report ROUGE scores \cite{lin-hovy-2003-automatic} with respect to the gold summary. 
For \mbox{\mranews{}(S-A)}, we select the first three sentences in each of the assisting documents and calculate ROUGE with respect to the gold summary individually; the reported ROUGE is then averaged over the individual documents.
Furthermore, we use the \emph{extractive oracle} (\oracle), which is often used
 as an upper bound for extractive models \cite{10.5555/3298483.3298681,narayan-etal-2018-ranking}. 
It creates an oracle summary by selecting the best possible set of sentences in the document that gives the highest ROUGE score with respect to the gold summary.\footnote{We use the greedy method from \url{https://github.com/pltrdy/extoracle_summarization}.}
For \mranews{}(S-D), we select the best three sentences in the main document as the summary, while for \mranews{}(S-A), we choose the best three sentences from all assisting documents as the summary. 
All selected summaries are evaluated using ROUGE against gold summaries.
Higher ROUGE scores intuitively correspond to more extractive summaries. 

The middle and right sections in Table~\ref{table:data_novelty} show the \lead{} and \oracle{} results, respectively. 
Both 
reach high scores on \mranews{}(S-D), while \oracle{} shows that improved content selection helps more. 
Although both methods achieve a much worse performance on \mranews{}(S-A) compared to  \mranews{}(S-D), ROUGE scores are comparable to the ones reached on XSum.
This confirms the conclusions we draw from the n-grams novelty metric. 


\subsection{Informativeness of Assisting Documents}
\label{subsec:informativeness}

Next, we evaluate the informativeness of the assisting documents with the following four metrics:
We use n-gram novelty and \textsc{Ext-Oracle} from the previous section
for measuring extractive token overlap. We also introduce two new metrics based on semantic similarity, which abstracts away from the actual tokens and is thus better suited for abstractive summarization.

\noindent $\bullet$ {\bf N-gram novelty} 
\mranews{}(S-D\&A) in Table~\ref{table:data_novelty} reports the n-gram novelty of the summaries with respect to their main and assisting documents, which is substantially lower than \mranews{}(S-D). 
Introducing the assisting documents 
contributes new
information to support the summary better.

\noindent $\bullet$ {\bf \oracle{}} 
\mranews{}(S-D\&A) in Table~\ref{table:data_novelty} contains the best three sentences from the main and assisting documents against the summary. 
The higher ROUGE scores on \mranews{}(S-D\&A), as compared to \mranews{}(S-D), indicate that assisting documents $A$ contribute additional information to the summaries, which is absent from the main document $D$.

\noindent $\bullet$ {\bf Summary Fact-weights} evaluate the semantic correspondence between a document and its summary using a 
representation based on 
``{\em facts}''. 
We follow \citet{xu-etal-2020-fact} and represent facts in a sentence by adapting Semantic Role Labelling \cite{palmer-etal-2005-proposition}, which roughly captures ``who did what to whom'' in terms of predicates and their arguments. 
The facts in the document and summary are represented as $\left \{ F^D_1,  F^D_2, \cdots F^D_I \right \}$ and $\left \{ F^S_1, F^S_2, \cdots F^S_J \right \}$, respectively. 
We apply automatic content weighting as defined in \cite{xu-etal-2020-fact} 
 and weight each fact $F_j$ in the summary using its maximum semantic similarity to the facts in the document $w^f_j = \max_{i \in I} d_{ij}^f$, where
$d_{ij}^f$ is the semantic similarity based on BERT embeddings \cite{devlin-etal-2019-bert}. 
The \emph{Summary Fact-weights} score is then defined as the average weights over all facts in the summary:
\begin{equation}\label{Eq:0}
\small
\textit{SFweights} = \textup{avg}_{j=1\cdots J}w^f_j  \in [-1,1]
\end{equation}
A high $\textit{SFweights}$ score indicates that the facts in the summaries are well supported by the facts mentioned in the documents.

The top section in Table~\ref{table:data_factual} shows \textit{SFweights} scores reported on \mbox{\mranews{}(S-D)}, \mranews{}(S-A) and \mranews{}(S-D\&A), which weight facts in the summaries using facts in the main document, assisting documents, and both, respectively. 
As expected, \textit{SFweights} on \mranews{}(S-D) is higher than on \mranews{}(S-A), indicating that the summary mainly contains facts from the main document $D$ and can't be generated from assisting documents alone.
However, \textit{SFweights} on \mbox{\mranews{}(S-D\&A)} is higher than on \mbox{\mranews{}(S-D)}, which indicates that the assisting documents provide additional information beyond the main document and still preserve the facts in the summary.

\noindent $\bullet$ {\bf Assist Rate}  extends 
\textit{SFweights} by first weighting the facts in the summary using the main document $\left [ w_1^{fc}, w_2^{fc}, \cdots w_J^{fc} \right ]$, and the assisting document $\left [ w_1^{fa}, w_2^{fa}, \cdots w_J^{fa} \right ]$.
It is then defined as: 
\begin{align}\label{Eq:1}
\small
\textit{AsstRate} &= \frac{\sum_{j=1}^{J}f\left ( w^{fc}_j, w^{fa}_j \right )}{J} \\
  f\left ( w^{fc}_j, w^{fa}_j \right )&=\begin{cases}
    1, & \text{if $w^{fa}_j > w^{fc}_j$}.\\
    0, & \text{otherwise}.
  \end{cases}
\end{align}
where $J$ is the number of facts in the summary.
\mbox{\textit{AsstRate}} represents the percentage of the facts in the summary that are \textit{better represented} in the assisting documents than in the main document.\footnote{While the main document might contain the facts, their structure is more accurately covered 
in assisting documents.}
We also extend the fact-level \textit{AsstRate} to the summary level, where we report the proportion of summaries in the entire corpus whose fact-level \textit{AsstRate} is over 0.
The bottom section in Table~\ref{table:data_factual} shows that more than 27\% of facts existing in 30\% of summaries are better grounded on assisting documents.

\begin{table}
  \centering\small
  \begin{tabular}{l|l} 
  \hline
  {\bf Metrics} & {\bf Results} \\ \hline \hline
  SFweights MiRA(S-D)           & 0.633  \\
  SFweights MiRA(S-A)           & 0.584  \\
  SFweights MiRA(S-D\&A)        & 0.658  \\ \hdashline
  AsstRate [fact level] (\%)~    & 27.67  \\
  AsstRate [summary level] (\%)~ & 30.20  \\
  \hline
  \end{tabular}
  \caption{Summary Fact-weights (\textit{SFweights}) and Assist Rate (\textit{AsstRate}) show that the assisting documents provide additional information beyond the main document to the summary. \label{table:data_factual}}
\end{table}

\section{Benchmarks}

\subsection{Baselines}\label{subsec:baselins}
After establishing the lower and upper bounds for extractive summarization models (see Section~\ref{subsec:ext_bias}), we mainly focus on abstractive approaches in our experiments. 
Many existing powerful sequence to sequence models, e.g.\ BART \cite{lewis-etal-2020-bart}, target conditional text generation tasks including summarization. 
Specific instances of Transformer-based \cite{46201} models, such as Longformer \cite{Beltagy2020Longformer}, BigBird \cite{49533}, PEGASUS \cite{pmlr-v119-zhang20ae}, HEPOS \cite{DBLP:journals/corr/abs-2104-02112} and Hierarchical Transformer (HT) \cite{liu-lapata-2019-hierarchical}, are designed for encoding long sequences.

In order to measure the effect that transfer learning has on \mranews{}, we try 
BART-large\footnote{
  Implementation used: \url{https://huggingface.co/transformers/model_doc/bart.html}.
  } \cite{lewis-etal-2020-bart} which is pre-trained and can take 1024 words as input, and HT\footnote{
  We use the implementation from \url{https://github.com/nlpyang/hiersumm}.
  } 
\cite{liu-lapata-2019-hierarchical} 
which is trained from scratch and can handle a longer input of up to 2000 words. We test four different variants for both models:

\noindent $\bullet$ {{\bf Single (-S):}} We only consider the main document as the input for generating the summary, replicating the SDS setup. 

\noindent $\bullet$ {{\bf Concatenation (-C):}} We simply append the assisting documents at the end of the main document. Since each document contains around 700 words on average (see Table~\ref{table:data_length}), we truncate the main document to half the size of the model capacity, i.e. 500 words for BART-large and 1000 words for HT, respectively. To include information from all assisting documents, we truncate each of them to fill the remaining half of the model capacity evenly. 

\noindent $\bullet$ {{\bf Pipeline (-P):}} Previous approaches T-DMCA \cite{j.2018generating}, TLM \cite{pilault-etal-2020-extractive} and SEAL \cite{DBLP:journals/corr/abs-2006-10213} show that 
long input settings for abstractive summarization benefit from a content extraction preprocessing step.  
%
We thus introduce a simple weakly supervised content extraction method for the assisting documents, and concatenate the selected content to the end of the main document on the input. Note that the content selection in \mranews{} is conditioned on the main document, which is different from content selection in both SDS and MDS that select sentences without additional conditioning.

In particular, we first compute a contextual embedding for each sentence in both main and assisting documents using BERT \cite{devlin-etal-2019-bert}, represented as $D^{emb} = \left \{ e^D_1, e^D_1, \cdots e^D_N \right \}$ and $A^{emb} = \left \{ e^A_1, e^A_1, \cdots e^A_K \right \}$. Then we calculate the semantic relevance for each sentence in the assisting documents with respect to each sentence in the main document, as the cosine distance between their sentence embeddings. In turn, we select the sentence $k$ in the assisting documents if:
\begin{equation*}\label{Eq:4}
\begin{split}
& \alpha_1 < \textup{avg}_{n=1:N}\textup{cosdist}\left ( e_n^D, e_k^A \right ) < \beta_1, \text{and}  \\
& \alpha_2 < \textup{max}_{n=1:N}\textup{cosdist}\left ( e_n^D, e_k^A \right ) < \beta_2, \text{and} \\
& \alpha_3 < \textup{min}_{n=1:N}\textup{cosdist}\left ( e_n^D, e_k^A \right ) < \beta_3 .
\end{split}
\end{equation*}
All thresholds are calculated on the training set using the gold content selection introduced in the following variant.\footnote{
We calculate the avg. cosdist(), max. cosdist() and min. cosdist() for each sentence in the gold content selection with respect to the corresponding main document. Then we calculate the distribution of the scores in each of these three category in terms of mean $\mu$ and variance $\sigma$. The lower and upper bound thresholds in each category are ($\mu-\sigma$) and ($\mu+\sigma$). Hence we get $\alpha_1$=0.73, $\beta_1$=0.83, $\alpha_2$=0.81, $\beta_2$=0.91, $\alpha_3$=0.59, $\beta_3$=0.75.
}

\noindent $\bullet$ {{\bf{Gold (-G):}}} 
We introduce a 
``heuristic'' upper bound baseline by replacing the weakly supervised procedure above with gold content selection, following a procedure introduced by \cite{pilault-etal-2020-extractive, 10.5555/3298483.3298681}
%
We select top sentences $s_D$ from both main and assisting documents 
 based on their extraction scores computed against sentences $s_S$ from the ground-truth summary $S$:  $\textup{SCORE}_{ext}(s_D) = \frac{1}{3}\sum_{r\in 1, 2, L}\textup{ROUGE}_r\left ( s_D, s_S \right )$, where $s_D \in D\cup\textbf{A}; s_S \in S$.
 We clean up the sentences that are selected multiple times.

\subsection{Evaluation Metrics}

We evaluate the approaches described in Section~\ref{subsec:baselins} from four perspectives:
 
\noindent $\bullet$ {{\bf Similarity to Reference}} focuses on evaluating the generated summary with respect to its similarity to a human-authored ground-truth reference summary.
We adopt the exact-matching metric {\bf {\em ROUGE}} \cite{lin-hovy-2003-automatic} and the soft-matching metric {\bf {\em BertScore}} \cite{bert-score}.

\noindent $\bullet$ {\bf Extractiveness level} aims at the bias of each system towards generating extractive summaries. We introduce the {\bf {\em n-grams coverage}}, which equals to $1-\text{n-gram novelty}$ (see Section~\ref{sec:data_analysis}), to measure the percentage of n-grams in the generated summary that appear in the main and assisting documents. Higher n-gram coverage scores indicate that the system is more extractive.

\noindent $\bullet$ {{\bf Support from Assisting Documents}} measures the proportion of information appearing in the generated summary that originates from assisting documents only. We propose the {\bf {\em n-grams coverage}} over  n-grams in the generated summary with respect to the n-grams that appear {\bf only} in the assisting documents (i.e, not in the main document). 

\noindent $\bullet$ {{\bf Extrinsic Hallucination}} aims at evaluating how much the facts in the generated summary are grounded in the main and the assisting documents. We adopt the {\bf {\em SFweights}} introduced in Section~\ref{subsec:informativeness}. A high SFweights score indicates that the facts in the generated summary are unlikely to be a result of extrinsic hallucination.

\section{Experiment results}

\begin{table}[tb]
  \centering\small
  \begin{tabular}{l|p{0.6cm}p{0.6cm}p{0.8cm}|p{0.3cm}p{0.5cm}p{0.5cm}} 
  \hline
  \multirow{2}{*}{{\bf Systems}} & \multicolumn{3}{c|}{{\bf ROUGE}} & \multicolumn{3}{c}{{\bf BertScore}}  \\
    & {\bf R1} & {\bf R2} & {\bf RL} & P & R & F1  \\ \hline \hline
  BART-S & 46.07 & 34.19 & 42.14 & .701 & .674 & .684 \\
  BART-C & 45.44 & 33.70 & 41.56 & .701 & .666 & .679  \\
  BART-P & {\bf 46.32} & {\bf 34.31} & {\bf 42.29} & .701 & {\bf .677} & {\bf .685} \\
  \hdashline
  HT-S & 46.76 & {\bf 36.18} & {\bf 43.22} & .685 & .682 & .680  \\
  HT-C & 46.77 & 36.06 & 43.11 & {\bf.690} & .682 & {\bf.682}   \\ 
  HT-P & {\bf 46.83} & 36.08 & 43.13 & .684 & {\bf .686} & .681  \\ \hline
  BART-G & 60.09 & 46.72 & 55.39 & .769 & .745 & .755   \\
  HT-G & 55.16 & 43.15 & 51.02 & .716 & .731 & .721  \\\hline
  \end{tabular}
  \caption{Evaluation on ROUGE and BertScore. 
  \label{table:res_1}}
\end{table}

\begin{table*}[tb]
  \centering\small
  \begin{tabular}{l|cccc|cccc|c} 
  \hline
  \multirow{2}{*}{{\bf Systems}} & \multicolumn{4}{c|}{{\bf Extractiveness level (\%)}} & \multicolumn{4}{c|}{{\bf Support from Assisting Documents (\%)}} & \multirow{2}{*}{{\bf SFweights}}  \\
    & {\bf 1-gram} & {\bf 2-gram} & {\bf 3-gram} & {\bf 4-gram} &  {\bf 1-gram} & {\bf 2-gram} & {\bf 3-gram} & {\bf 4-gram} \\ \hline \hline
    BART-S & 87.24 & 72.94 & 63.85 & 57.61 & 1.76 & 1.32 & 0.55 & 0.24 & .814 \\
    BART-C & 88.37 & 75.74 & 66.98 & 60.71 & 2.99 & 3.22 & 2.24 & 1.62 & {\bf .860} \\
    BART-P & 87.79 & 74.16 & 65.19 & 59.00 & 2.57 & 2.37 & 1.39 & 0.90 & .850 \\
    \hdashline
    
    HT-S & 98.14 & 95.70 & 93.98 & 92.82 & 0.51 & 0.38 & 0.16 & 0.08 & .840 \\
    HT-C & 99.48 & 98.46 & 97.58 & 96.86 & 1.53 & 2.37 & 2.33 & 2.19 & {\bf .881} \\ 
    HT-P & 99.20 & 97.86 & 96.83 & 96.00 & 0.92 & 1.18 & 1.01 & 0.87 & .860 \\ \hline
  
    BART-G & 87.14 & 71.42 & 60.94 & 53.91 & 4.22 & 4.96 & 3.59 & 2.65 & .817  \\
    HT-G & 98.88 & 96.57 & 94.58 & 93.10 & 2.82 & 4.56 & 4.48 & 4.17 & .845 \\ \hline
  \end{tabular}
  \caption{Evaluation of extractiveness level using {\bf n-gram coverage} (left), support from assisting documents (middle) calculated by {\bf n-gram coverage} with respect to the n-grams appearing in the assisting documents only, and extrinsic hallucination (right) evaluated using SFweights. \label{table:res_2}}
 \end{table*}

\begin{figure*}[tb]
  \includegraphics[width=1.0\textwidth]{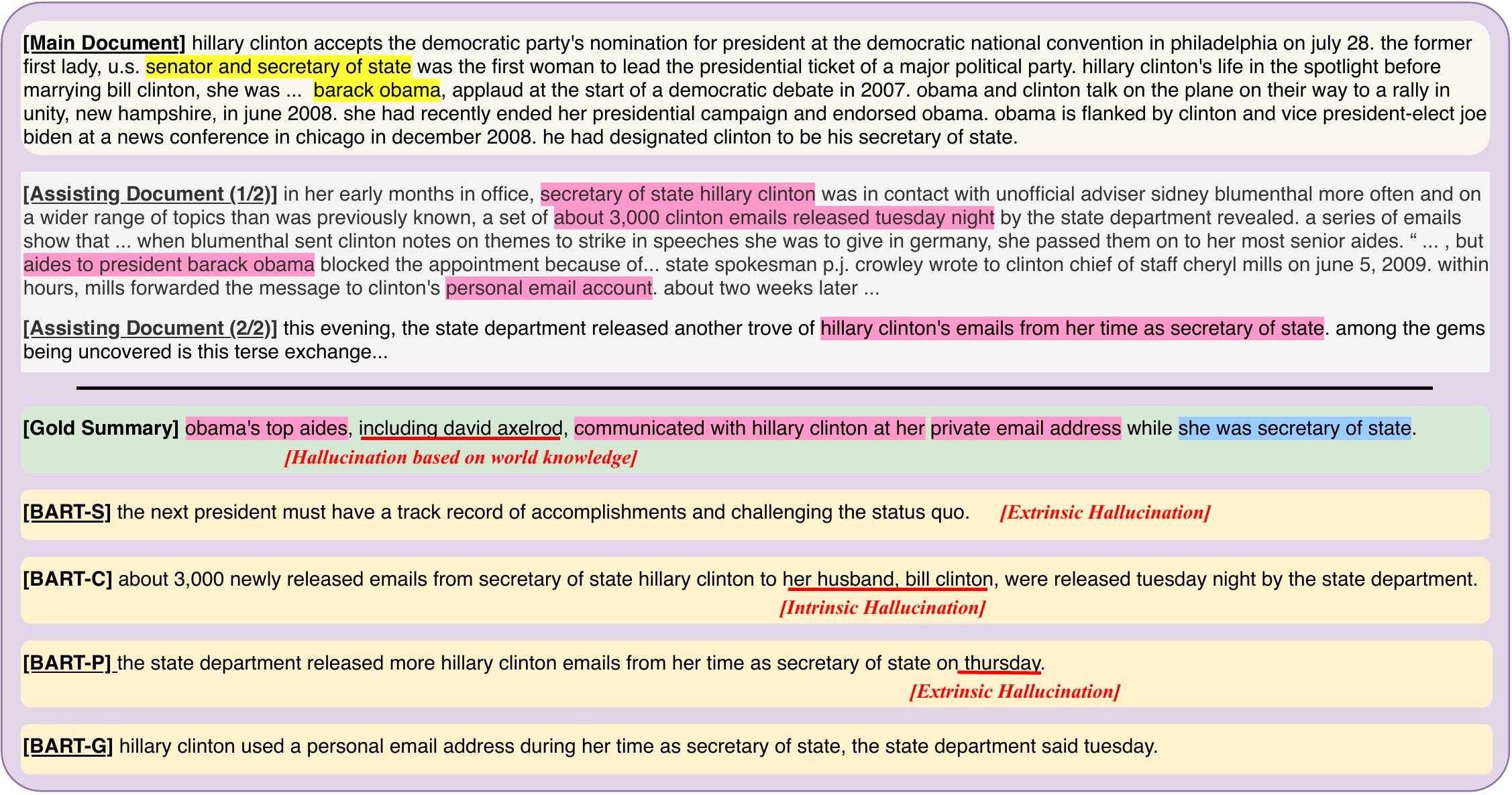}
  \caption{An example from \mranews, where the key information in the gold summary and summaries generated by systems conditioning on the main document (BART-S) or both on the main and assisting documents (rest variants) were only mentioned in the assisting documents. Facts in the gold summary supported by the assisting documents only are \colorbox{magenta!50!}{highlighted in pink}. Information grounded in both main and assisting documents is  \colorbox{cyan!70!}{highlighted in blue}. Other error type examples, including {\em Extrinsic Hallucination}, {\em World Knowledge-based Extrinsic Hallucination} and {\em Intrinsic Hallucination} in summaries are \textcolor{red}{{\bf {\em [labeled in red]}}}.}
  \label{fig:error_analysis}
  \vspace{-8pt}
\end{figure*}

\paragraph{Similarity to Reference.}
The results of reference-based automatic metrics are shown in Table~\ref{table:res_1}. 
The performance of BART and HT are comparable in most of the variants, which indicates that  systems trained from scratch on  \mranews{} are able to achieve similar performance to the systems fine-tuned on the pre-trained checkpoints. 

On most metrics, 
the concatenation variants (-C) of the models perform worse than the pipeline approaches (-P) and SDS-trained systems (-S).
On the other hand, both -P outperform the -S systems in most cases.
The gold systems (-G) achieve the best performance with a large margin. The performance of BART-G is even comparable with the upper bound of the extractive models (\oracle{} generated from \mranews{}(S-D\&A)). 
Hence, we 
conclude that (1) introducing assisting documents benefits the abstractive summary generation; (2) better content selection improves the performance of the abstractive models; (3) the margin between the gold upper baseline and the rest is notable, 
which suggests that there is 
room for improvement for content selection. 

\paragraph{Extractiveness Level.}
The results are shown in the left section of Table~\ref{table:res_2}. 
N-grams coverage scores for HT are much higher than BART's, with 4-grams over 90\%. This indicates that HT tends to generate very extractive outputs.
For each of the two models, the concatenation systems are more extractive than single-document and pipeline systems. For the BART variants, the gold system leans to generate more abstractive summaries compared to the remaining variants; for HT, the gold system is as extractive as all other variants.

\paragraph{Support from Assisting Documents.}
The middle section of Table~\ref{table:res_2} shows the amount of information each system learns  from the assisting documents alone. In both models, the gold, concatenate and pipeline variants include substantially more expressions occurring in the assisting documents compared to the single-document systems. 

\paragraph{Extrinsic Hallucination.}
The results in the right section of Table~\ref{table:res_2} show that
HT achieves a higher SFweights score, i.e. lower level of extrinsic hallucination, than BART 
-- probably due to the high extractiveness of HT. 
In other words, extractive summaries that copy sentences directly from the document tend to maintain higher SFweights scores. On the other hand, BART systems demonstrate a much higher level of abstractiveness, while preserving a similar SFweights score with HT. 
Thus, the BART systems do not introduce more hallucinations while generating abstractive summaries. 

Within each of the two models, summaries generated by each variant preserves a roughly similar level of extractiveness. 
In both models, concatenation and pipeline systems achieve a lower level of extrinsic hallucination compared to the single-document systems. SFweights for BART-G is lower than most other setups, probably due to a high level of abstractiveness in this system. 
To better understand the relation between introducing assisting documents and reducing extrinsic hallucinations, we conduct an example-based analysis in the next section.

\section{Hallucination Analysis}\label{sec:error-analysis}

We manually identify 4 types of hallucinations from a small random sample (30 main/assisting documents and summaries) from the development set of \mranews, as summarised in Table \ref{tab:hallucination_analysis}.
In particular, we examined claims in the summaries that were not mentioned in the main or assisting documents and were (1) erroneous ({\em Extrinsic Hallucinations}), (2) factual possibly due to pretraining ({\em World knowledge}), (3) only mentioned in the assisting document correctly ({\em Grounded Asst.}), or (4) mentioned in the main document in a different way ({\em Intrinsic}). 
We omit the HT variants from our analysis as their output is more extractive, and therefore less prone to hallucinations. 
The SDS variant of BART (BART-S) has the highest percentages of extrinsic (7) and intrinsic (4) hallucinations and a number of claims that are based on world knowledge 
(3). On the other hand, the inclusion of assisting documents sees an overall reduction in both types with up to 55\% on extrinsic hallucinations when using the assisting documents for training efficiently (BART-G). At the same time, we observe `extrinsic hallucinations' that are correctly grounded only on the assisting documents (11), and rarely \textit{guessed} based on pre-training 
(only 1 fact based on world knowledge). Interestingly, we also observed a number of facts (10) in the gold summary that are grounded exclusively on the assisting documents, further supporting the value of our approach. An example of outputs from variants of BART is shown in Figure~\ref{fig:error_analysis}.

    
  

\begin{table}[]
    \centering\small
    \begin{tabular}{l|c|c|c|c}\hline
    \textbf{Systems} &\textbf{Extr.} & \textbf{World} & \textbf{Asst.} & \textbf{Intr.} \\ \hline \hline
    GOLD & 1 & \textbf{10} & 11 & 0 \\ \hdashline
     BART-S & \textbf{7} & 3 & 0 & \textbf{4} \\
    BART-C & 0 & 0 & 6 & 2 \\
    BART-P & 3 & 1 & 3 & 1 \\
    \hdashline
    
  
    BART-G & 3 & 0 & \textbf{11} & 2 \\ \hline
    \end{tabular}
    \caption{Manual analysis of types of hallucinations (counterfactual extrinsic [\textbf{Extr.}], factual extrinsic based on world knowledge [\textbf{World}], grounded exclusively on assisting documents [\textbf{Asst.}], intrinsic [\textbf{Intr.}]) on a sample of 30 summaries from \mranews.}
    \label{tab:hallucination_analysis}
\end{table}

\section{Related Work}

\paragraph{Single Document Summarization} aims to
compress a single textual document while keeping salient information. 
SDS includes two directions: extractive summarization \cite{10.5555/3298483.3298681}
which aims at extracting salient sentences from the input document, and abstractive summarization \cite{46111, narayan-etal-2018-dont,DBLP:conf/nips/YangDYCSL19,
liu-lapata-2019-text,liu2020roberta,rothe-etal-2020-leveraging,JMLR:v21:20-074} which generates a novel short representation of the input. 

\paragraph{Multi-Document Summarization}
aims to compress multiple textual documents to a shorter summary \cite{fabbri-etal-2019-multi}. 
Approaches mainly focus on increasing the capacity of the encoder to process longer inputs \cite{liu-lapata-2019-hierarchical, Beltagy2020Longformer, 49533, pmlr-v119-zhang20ae, DBLP:journals/corr/abs-2104-02112}, leveraging knowledge graphs \cite{fan-etal-2019-using, li-etal-2020-leveraging-graph, jin-etal-2020-multi}, and including content selection steps \cite{nayeem-etal-2018-abstractive, wang-etal-2020-spectral, xu-lapata-2020-coarse, grenander-etal-2019-countering, j.2018generating}.

\paragraph{Hallucinations in Summarization} 
are a well established problem \cite{maynez-etal-2020-faithfulness,AAAI1816121, falke-etal-2019-ranking}.
Previous research aimed to reduce hallucination by adapting model architectures, training and decoding, e.g.\ \citet{AAAI1816121, zhang-etal-2020-optimizing, falke-etal-2019-ranking, zhao-etal-2020-reducing}. However, we are the first research aiming to reduce the hallucinations by adapting the dataset.

\section{Conclusions and Future Work}

In this work, we found that up to 36\% facts in the ground truth summaries in traditional SDS datasets are not faithful to the source article. In other words, the ground truth summaries also  contain `extrinsic hallucinations'.
Summarization models trained on such data will be prone to extrinsic hallucinations. 
To tackle this problem, we introduce a new task, Multi-Resource-Assisted News summarization, which produces a summary based on the events present in the main article while reaching to a set of assisting documents for complementary background. 
We release the \mranews dataset, which includes multiple assisting news articles from different news resources for each document-summary pair. 
Our newly introduced evaluation metrics confirm that introducing assisting documents offers better grounding to more than 27\% facts in the reference summaries.
We report benchmark results on \mranews{}. We also show that the model trained with assisting documents produces 55\% less counterfactual hallucinations than a model trained only with main documents. 

In future work, we plan to explore a retrieval-based approaches \cite{azzopardi2012incremental, bouras2012clustering} that are able to search and filter relevant assisting documents for a given news event, without the help of human-edited resources such as \url{newser.com}. 
In the paper, we demonstrated that the assisting documents contain useful facts to support the summarization of the main news event. Thus, efficient content selection that eliminates noise and grounds in the relevant facts appearing in either main or assisting documents will also be explored in our future work.

\section*{Acknowledgments}
This research received funding from the EPSRC project AISec (EP/T026952/1), Charles University project PRIMUS/19/SCI/10, a Royal Society research grant (RGS/R1/201482), a Carnegie Trust incentive grant (RIG009861). This research also received funding from Apple to support research at Heriot-Watt University and Charles University. We thank the anonymous reviewers and the area chair for their helpful comments and hard work.

\bibliography{custom}
\bibliographystyle{acl_natbib}

\clearpage
\onecolumn
\appendix


\end{document}